\def\year{2021}\relax
\documentclass[letterpaper]{article} 
\usepackage{aaai21}  
\usepackage{times}  
\usepackage{helvet} 
\usepackage{courier}  
\usepackage[hyphens]{url}  
\usepackage{graphicx} 
\usepackage{natbib}  
\usepackage{caption} 
\usepackage{amsmath}
\usepackage{amssymb}
\usepackage{booktabs}
\usepackage{multirow}
\usepackage{array}

\usepackage[switch]{lineno}

\title{Building Interpretable Interaction Trees for Deep NLP Models}
\author{
    Die Zhang,
    Huilin Zhou,
    Hao Zhang,
    Xiaoyi Bao, \\
    Da Huo,
    Ruizhao Chen,
    Xu Cheng,
    Mengyue Wu,
    Quanshi Zhang\thanks{Corresponding author. This work was done under the supervision of Dr. Quanshi Zhang.} \\
}
\affiliations{

    Shanghai Jiao Tong University \\
    \{zizhan52, zhouhuilin116, 1603023-zh, zjbaoxiaoyi\}@sjtu.edu.cn \\
    \{sjtuhuoda, stelledge, xcheng8, mengyuewu, zqs1022\}@sjtu.edu.cn \\

}
\begin{document}
	
\maketitle
\begin{abstract}
This paper proposes a method to disentangle and quantify interactions among words that are encoded inside a DNN for natural language processing. We construct a tree to encode salient interactions extracted by the DNN. Six metrics are proposed to analyze properties of interactions between constituents in a sentence. The interaction is defined based on Shapley values of words, which are considered as an unbiased estimation of word contributions to the network prediction. Our method is used to quantify word interactions encoded inside the BERT, ELMo, LSTM, CNN, and Transformer networks. Experimental results have provided a new perspective to understand these DNNs, and have demonstrated the effectiveness of our method.
\end{abstract}

\section{Introduction}
Deep neural networks (DNNs) have shown promise in various tasks of natural language processing (NLP), but a DNN is usually considered as a black-box model. In recent years, explaining features encoded inside a DNN has become an emerging direction. Based on the inherent hierarchical structure of natural language, many methods use latent tree structures of language to guide the DNN to learn interpretable feature representations~\cite{choi2018learning,drozdov2019unsupervised,shen2018neural,shen2018ordered,shi2018tree,
	tai2015improved,wang2019tree,yogatama2016learning}. However, the interpretability usually conflicts with the discrimination power~\cite{bau2017network}. There is a considerable gap between pursuing the interpretability of features and pursuing superior performance.

Therefore, in this study, we aim to explain a trained black-box DNN in a post-hoc manner, so that the explanation of the DNN does not affect its performance. This is essentially different from previous studies of designing new network architectures or losses to learn interpretable features, \emph{e.g.}~physically embedding tree structures into a DNN.

Given a trained DNN, in this paper, we propose to analyze interactions among input words, which are used by the DNN to make a prediction.
Our method generates a tree structure to objectively reflect interactions among words.
Mathematically, the interaction of several words is quantified as the difference of the contribution between the case when these words  contribute jointly to the prediction and the case when each individual word contributes independently to the prediction.
The interaction between words may bring either positive or negative effects on the prediction. For example, the word \emph{green} and the word \emph{hand} in the sentence \emph{he is a green hand} have a strong and positive interaction to the prediction of the person's identity, because the words \emph{green} and \emph{hand} indicate a ``novice" jointly, rather than work individually to represent a hand with a green color.

The core challenge in this study is to guarantee the objectiveness of the explanation. \emph{I.e.} the tree needs to reflect true interactions among words without significant bias.
The Shapley value is widely considered as a unique unbiased estimation of the word contribution~\cite{lundberg2017unified}, which satisfies four desirable properties, \emph{i.e.~linearity, dummy, symmetry and efficiency}~\cite{Grabisch1999AnAA}.
Thus, we define the interaction benefit among words based on the Shapley value.
Let us consider a constituent with $m$ words. $\phi_{1}, \phi_{2},\dots, \phi_{m}$ denote numerical contributions of each word to the prediction of a DNN, respectively.
$\phi_{all}$ represents the numerical contribution of the entire constituent to the prediction.
Hence, $ B =  \phi_{all} - \sum_{i=1}^{m}{\phi_{i}}$ measures the interaction benefit of this constituent.
If $B > 0$, interactions among these $m$ words have positive effects on the prediction; otherwise, negative effects. Here, $\phi_1,...,\phi_m,\phi_{all}$ can be computed as Shapley values.

\begin{figure}[t]
	\centering
	\includegraphics[width=0.61\columnwidth]{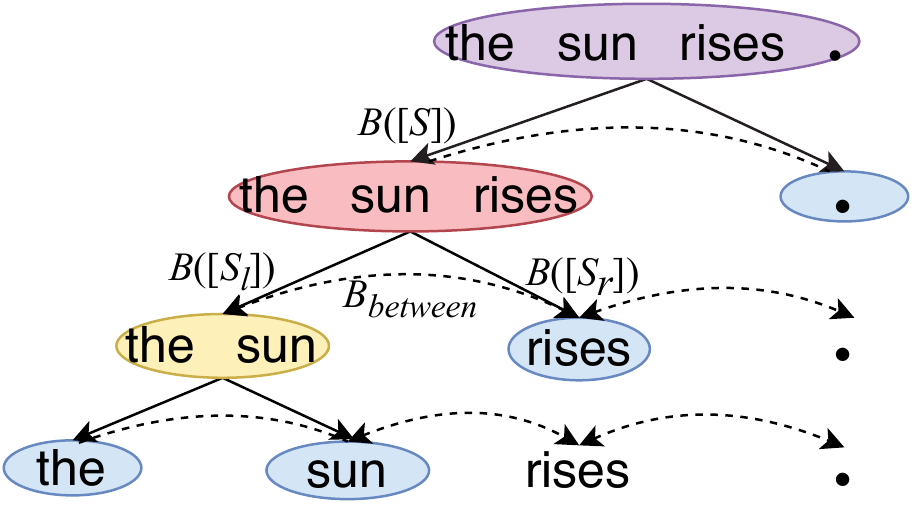}
	\caption{A tree to represent interactions among words. The tree is  built to explain a trained DNN. Each leaf node (blue) represents an input word in the sentence. Each non-leaf node encodes the significance of interactions within a constituent.}
	\label{tree}
\end{figure}

Given a trained DNN and an input sentence with $n$ words, Figure~\ref{tree} shows the tree structure that reflects word interactions encoded inside the DNN. In the tree, $n$ leaf nodes represent $n$ input words. Each non-leaf node corresponds to a constituent of the input sentence. A parent node connects two child nodes with significant interaction benefits.
We use the parent node to encode interactions among its child sub-constituents.
More specifically, there are two types of interactions among words, \emph{i.e.}~(1) interactions within a constituent and (2) interactions between constituents.

{$\bullet$} \textbf{Interactions within a constituent} exist among any two or more words in the constituent. For the sentence ``\emph{the sun is shining in the sky}," interactions within the constituent \emph{in the sky} consist of interactions among all combinations of words, including interactions (1) between (\emph{in}, \emph{the}), (2) between (\emph{the}, \emph{sky}), (3) between (\emph{in}, \emph{sky}) and (4) among (\emph{in}, \emph{the}, \emph{sky}).

{$\bullet$} \textbf{Interactions between constituents.} In
the aforementioned sentence, interactions between the constituent \emph{the sun} and its adjacent constituent \emph{is shining} are composed of all potential interactions among all combinations of words from the two constituents, including interactions between (1) (\emph{the}, \emph{is}), (2) (\emph{the}, \emph{shining}), (3) (\emph{sun}, \emph{is}), (4) (\emph{sun}, \emph{shining}), (5) (\emph{the}, \emph{is shining}), (6) (\emph{sun}, \emph{is shining}), (7) (\emph{the sun}, \emph{is}), (8) (\emph{the sun}, \emph{shining}), (9) (\emph{the sun}, \emph{is shining}).

The tree selects and encodes the most salient interactions among words, in order to reveal the signal processing in a DNN. We further propose additional metrics to diagnose interactions among words, \emph{e.g.}~the quantification of interactions within a constituent, the quantification of interactions between two adjacent constituents, and ratios of interactions that are modeled and unmodeled by the tree.

Theoretically, our method can be used as a generic tool to analyze various DNNs, including the BERT~\cite{devlin2018bert}, ELMo~\cite{peters2018deep}, LSTM~\cite{hochreiter1997long}, CNN~\cite{kim2014convolutional} and Transformer~\cite{vaswani2017attention}. Experimental results have demonstrated the effectiveness of our method.

\textbf{Contributions} of this paper can be summarized as follows. (1) We propose a method to extract and quantify interactions among words. (2) A tree structure is automatically generated to represent salient interactions encoded in a DNN. (3) We further design six metrics to analyze interactions, which provides new perspectives to understand DNNs.

\section{Related Work}

\subsubsection{Hierarchical representations of natural language.}
Many studies integrated hierarchical structures of natural language into DNNs for better representations~\cite{tai2015improved, dyer2016recurrent, wang2019self, wang2019tree}.
\citet{chung2016hierarchical} revised an RNN to learn the hierarchical structure of sequential data. \citet{shen2018ordered} designed a novel recurrent architecture to automatically capture the latent tree structure of an input sentence.
Other studies learned syntactic parsers~\cite{drozdov2019unsupervised,htut2019inducing, kitaev2018multilingual,li2019imitation, li2019specializing, mrini2019rethinking}, although these methods pursued a high parsing accuracy, instead of explaining the DNN.
Essentially, the learning of the syntactic parser aimed to make the parser fit syntactic structures defined by human experts. In contrast, we intend to provide a method to analyze DNNs in a post-hoc manner, without being affected by the subjective bias from humans.

\emph{Post-hoc explanations of DNNs:} Some studies measured the representation capacity to understand DNNs~\cite{guan2019towards,Cheng_2020_CVPR,Liang2020Knowledge}.
\citet{voita-etal-2019-bottom} studied how token representations changed from layer to layer. \citet{coenen2019visualizing, raganato-tiedemann-2018-analysis} exploited the attention weights of models to analyze syntactic and semantic information encoded in internal representations.
\citet{yogatama2018memory} evaluated the ability of various RNNs to capture syntactic dependencies. Another line of research was to estimate word importance to the prediction based on Shapley values~\cite{shapley1953value}, such as SHAP~\cite{lundberg2017unified}, L/C-Shapley~\cite{chen2018shapley}.
\citet{murdoch2018beyond} estimated contributions of input words to the prediction of an LSTM as well as inter-word relationships.\footnote{Although they called the inter-word relationships \textit{interactions}, such interactions had essential difference from our interactions.}
\citet{singh2018hierarchical} and \citet{jin2019towards} generated hierarchical explanations for word/phrase importance.

Unlike above studies of estimating attribution/saliency /contribution/importance of input units, we focus on interactions among words encoded inside DNNs.
\citet{janizek2020explaining} explained pairwise feature interactions by extending the Integrated Gradients explanation method.  \citet{10.1093/bioinformatics/bty575} identified interactions between all pairs of discrete features in an input DNA sequence.
\citet{cuilearning} estimated global pairwise interactions from a trained Bayesian neural network.
\citet{tsang2018detecting} detected statistical interactions from the weights of feedforward neural networks.
\citet{NIPS2018_7822} proposed to separate feature interactions based on regularization, and could only be applied to fully connected multilayer perceptrons.
\citet{lundberg2018consistent} defined SHAP interaction values to quantify interaction effects between two features.
\citet{chen2020generating} generated hierarchical explanations of DNNs based on the SHAP interaction value. \citet{chen2019ls} used a \emph{``predefined''} syntactic constituency structure to assign an importance score to each word, and to quantify interactions\footnote{The deviation of composition from linearity.} between sibling nodes on a parse tree. This study had considerable impacts, but it did not learn the linguistic structure.

However, these studies mainly focus on interactions between two variables~\cite{janizek2020explaining, 10.1093/bioinformatics/bty575, cuilearning, lundberg2018consistent,chen2020generating} or are limited to multilayer perceptron architectures~\cite{tsang2018detecting, NIPS2018_7822}. Instead, we aim to quantify interactions among multiple variables in DNNs with arbitrary architectures without any prior linguistic structure. More specifically, our method uses a tree to organize the extracted interactions hierarchically.

\subsubsection{Shapley values.} The Shapley value~\cite{shapley1953value} was first introduced in game theory.
Given a game with multiple players, each player is supposed to pursue a high award. Some players may form a coalition to pursue more awards. The Shapley value is widely considered as a unique unbiased approach to fairly allocating the total award of a coalition to each player (here, the award of each player is also termed the \emph{contribution} of this player).

Given a game $v$ with $n$ players, $N=\{1,2,...,n\}$, let $2^N=\{S|S\subseteq N\}$ denote all the potential subsets of $N$. $v\colon 2^N \mapsto \mathbb{R}$ is a set function mapping from each subset to a real number. For any subset of players $S \subseteq N$, $v(S)$ represents the score obtained by the set of players $S$. $v(\varnothing)$ represents the baseline score without any players. Thus, $v(S)-v(\varnothing)$ corresponds to the award obtained by players in $S$. Considering the player $a \notin S$, if player $a$ joins $S$, $v(S \cup \{a\}) - v(S)$ is considered as the marginal award/contribution of player $a$.
The Shapley value $\phi(a)$ is an unbiased estimation of numerical contribution of player $a$ in the game as follows.
\begin{equation}
\label{shapley}
{\phi}(a) =\!\!\!\!\!\! \sum\limits_{S \subseteq N\backslash \{a\}}\!\!\!{\frac{(|N|-|S|-1)!|S|!}{|N|!}}(v(S \cup \{a\}) - v(S))
\end{equation}

The fairness of Shapley values is ensured by the four following properties~\cite{weber1988probabilistic}:

{$\bullet$} Linearity property: If two games $v$ and $w$ are combined into a single game $v\!+\!w$, then the Shapley value of each player $a\!\in\!N$ can be added, \emph{i.e.}~${\phi}(a|v+w)\!\! =\!\! {\phi}(a|v) + {\phi}(a|w)$.

{$\bullet$} Dummy property: A dummy player  $a\!\in\! N$ satisfies $\forall S\!\! \subseteq\!\! N \setminus\{a\}$, $v(S \cup \{a\})\! =\! v(S) + v(\{a\})$. Then, ${\phi}(a)=v(\{a\})-v(\varnothing)$, \emph{i.e.} player $a$ has no interaction to any coalition.

{$\bullet$} Symmetry property:
Given two players $a, b\in N$, if $\forall S\! \subseteq\!\! N\setminus\{a,b\}$,
$v(S\cup \{a\}) = v(S\cup \{b\})$, then  $\phi(a) = \phi(b)$.

{$\bullet$} Efficiency property: The overall award can be distributed to all players, \emph{i.e.}~$\sum_{a\in N}{\phi}(a)= v(N)-v(\varnothing)$.

Due to the exponential number of sets in $N$, the computation of Shapley values is NP-hard.
A sampling-based method~\cite{castro2009polynomial} can be used to approximate Shapley values.

\section{Algorithm}

\subsection{Interactions in game theory}
\subsubsection{Interactions between two players.} In game theory, some players may interact with each other, and form a coalition to win a higher award. The interaction between two players is quantified as the additional award when the two players collaborate \emph{w.r.t.} when they play individually.
Considering that the Shapley value is an unbiased estimation of each player's award/contribution~\cite{lundberg2017unified}, we quantify interactions based on the Shapley value. Suppose that there are $n$ players $N = \{1, 2,..., n\}$ in a game $v$.
Without loss of generality, we randomly select a pair of players $a, b \in N$. Shapley values of players $a$ and $b$ are denoted by $\phi(a)$ and $\phi(b)$, respectively. If players $a$ and $b$ cooperate to form a coalition $S_{ab} = \{a, b\}$, we can consider this coalition as a new singleton player, which is represented using brackets, $[S_{ab}]$. \textit{In this way, the game can be considered to have $n-1$ players, and one of them is the singleton player $[S_{ab}]$.} \emph{I.e.}~$a$ and $b$ always appear together in the game.
The interaction benefit between $a$ and $b$ is defined as $B([S_{ab}]) = \phi^{N\setminus \{a,b\} \cup \{[S_{ab}]\} }([S_{ab}]) - (\phi^{N\setminus \{b\}}(a) + \phi^{N\setminus \{a\}}(b))$.
{$N\setminus \{a,b\} \cup \{[S_{ab}]\}$} represents the set of players in $N$ excluding $a, b$ and being added a new singleton player $[S_{ab}]$.
The absolute value of the interaction benefit $|B([S_{ab}])|$
represents the significance of the interaction. $B([S_{ab}]) > 0$ indicates a cooperative relationship between players $a$ and $b$. Whereas, $B([S_{ab}]) < 0$ indicates an adversarial relationship between players $a$ and $b$.

\subsubsection{Extension to interactions among multiple players.}
We extend the two-player interaction to interactions among multiple players.
When the game has $n$ players, let us consider a subset of players $S\subsetneq N$ as a coalition, which is regarded as a new singleton player $[S]$.
The interaction benefit of the coalition $S$ is defined as follows.
\begin{equation}
B([S]) = {\phi}^{(N\backslash S)\cup \{[S] \}}([S]) - \sum\nolimits_{a\in S}{\phi}^{(N\backslash S)\cup \{a\}}(a)
\end{equation}
In this way, the interaction benefit measures the additional award/contribution brought by the singleton player $[S]$ \emph{w.r.t.} the individual award/contribution of each player computed in  Equation~\eqref{shapley} without requiring all players in $S$ to appear together.
The Shapley value ${\phi}^{(N\backslash S)\cup \{[S] \}} ([S])$ is computed only considering the set of players when we remove all players in $S$ from $N$ and add a new singleton player $[S]$ in the game. Similarly, ${\phi}^{(N\backslash S)\cup \{a\}} (a)$ is computed only considering the set of players when we remove all players in $S$ from $N$ and add the player $a$.
If $B([S])$ is greater/less than 0, interactions of players in $S$ have positive/negative effects, revealing the cooperative/adversarial relationship among players.

Furthermore, players in $S$ can be divided into two disjoint subsets $S_1, S_2$ (\emph{i.e.}~$S_1 \cap S_2 = \varnothing, S_1 \cup S_2 = S$). Accordingly, the interaction benefit can be decomposed into three terms:
\begin{equation}
\label{12_between}
B([S]) = B([S_1]) + B([S_2]) + B_{\textit{between}}(S_1, S_2)
\end{equation}
The first and second terms $B([S_1])$ and $B([S_2])$ indicate interaction benefits among players within $S_1$ and $S_2$, respectively.
The third term $B_{\textit{between}}(S_1, S_2)$ indicates interaction benefits among players selected from both $S_1$ and $S_2$.
$B_{\textit{between}}(S_1, S_2)$ will be introduced in detail later.

\subsubsection{Properties of interaction benefits.} The overall interaction benefit, $B([S]), S\subseteq N$, can be decomposed into elementary interaction components $I^{N}(S)$. The elementary interaction component was originally proposed in~\cite{Grabisch1999AnAA}.
The elementary interaction component $I^{N}(S)$ measures the marginal benefit received from the coalition $[S]$, from which benefits of all potential smaller coalitions $S'\subsetneq S$ are removed.
For example, let $S=\{a, b, c\}$. Then, $I^{N}(S)$ measures interactions caused by $[S]=(a,b,c)$, and ignores all potential interactions caused by coalitions of $(a, b), (a, c), (b, c), (a), (b), (c)$. Therefore, the elementary interaction component is formulated as follows.
\begin{equation}
I^{N}(S) = I^{(N\setminus S)\cup \{[S] \}}([S]) -\!\!\!\! \sum\limits_{S' \subsetneq S, S' \neq \varnothing}\!\!\! I^{(N\setminus S)\cup S'}(S')
\end{equation}
In particular, for any singleton player $[S]$,
we have $I^{(N\setminus S)\cup \{[S]\}}([S])=\phi^{(N\setminus S)\cup \{[S]\}}([S])$. Thus, we can compute $I^{N}(S)$ via dynamic programming.
We prove that $B([S])$ can be decomposed into elementary interaction components (the supplementary material shows the proof).
\begin{equation}
\label{B_I}
B([S]) = \sum\nolimits_{S' \subseteq S, |S'|>1} I^{(N\backslash S) \cup S'}(S')
\end{equation}

\subsection{Fine-Grained analysis of interactions between two sets of players}
\label{B_between}
Interactions between two sets of players $B_{\textit{between}}(S_1, S_2)$ can be further decomposed into three parts $\psi^{\textit{inter}}$, $\psi^{\textit{intra}}_1$, $\psi^{\textit{intra}}_2$. Please see the supplementary material for the proof.
\begin{equation}
\label{eq_intra}
B_{\textit{between}}(S_1, S_2) =  \psi^{\textit{inter}} + \psi^{\textit{intra}}_{1}+ \psi^{\textit{intra}}_{2}
\end{equation}
where
\begin{equation}
	\begin{aligned}
	\psi^{\textit{inter}}\!\!&=\sum\nolimits_{L\subseteq S, L\not\subset S_1,L\not\subset S_2,|L|>1}I^{(N\setminus S)\cup L}(L) \\
	\psi^{\textit{intra}}_{1}\!\!&=\!\!\!\!\!\!\!\sum\limits_{L\subseteq S_1,|L|>1}\!\!\!\!I^{(N\setminus S)\cup L}(L)-\!\!
	\sum\limits_{L\subseteq S_1, |L|>1}\!\!\!\!I^{(N\setminus S_1)\cup L}(L) \\
	&= B([S_1])|_{N'=(N\setminus S_2)} - B([S_1]) \\
	\psi^{\textit{intra}}_{2}\!\!&=\!\!\!\!\!\!\!\sum\limits_{L\subseteq S_2,|L|>1}\!\!\!\!I^{(N\setminus S)\cup L}(L)-\!\! \sum\limits_{L\subseteq S_2,|L|>1}\!\!\!\!I^{(N\setminus S_2)\cup L}(L) \\
	&= B([S_2])|_{N'=(N\setminus S_1)} - B([S_2])
	\end{aligned}
\end{equation}
$\psi^{\textit{inter}}$ represents all potential interaction benefits caused by sets of players whose elements are selected from both $S_1$ and $S_2$.
$B([S_1])|_{N'=(N\setminus S_2)}$ denotes interaction benefits of the singleton player $[S_1]$, when the set of players in the game is $N'=(N\!\setminus\!S_2)$.
$\psi^{\textit{intra}}_1$ indicates the difference of internal interactions among players in $S_1$ \emph{w.r.t.} the absence and presence of players in  $S_2$.

\subsection{Interactions encoded inside a DNN}

We aim to analyze interactions among words, which are encoded inside a trained DNN. Given an input sentence with $n$ words, we construct a tree to disentangle and quantify interactions among input words.
We first introduce Shapley values of input words \emph{w.r.t.} the prediction of the DNN. Here, we consider each word as a player, and the scalar output of a DNN as the aforementioned score $v$ in the game.
If a DNN has a scalar output, we can take the scalar output as the score $v$. If the DNN outputs a vector for multi-category classification, we select the score before the softmax layer corresponding to the predicted class as the score.
To compute $v(S)$, we mask words in $N\!\setminus\! S$ in the input sentence, and feed the modified input into the DNN. The embedding of the masked word is set to a dummy vector, which refers to a padding of the input to the DNN. Then, the Shapley value of each word/player $a$ is approximated using a sampling-based method~\cite{castro2009polynomial}.

As Figure~\ref{tree} shows, we construct a binary tree with $n$ leaf nodes. Each leaf node represents a word, while each non-leaf node represents a constituent.
Two adjacent nodes with strong interactions will be merged into a node in the next layer.
For each sub-structure of a parent node $S$ with two child nodes $S_l$ and $S_r$, according to Equation~\eqref{12_between}, $B([S])$ can be recursively decomposed into the sum of interaction benefits between two child nodes of all non-leaf nodes.
Please see the supplementary material for the proof.
\begin{equation}
\label{tree_equ}
	\begin{aligned}
	&B([S])=B([S_l]) + B([S_r]) + B_{\textit{between}}(S_l, S_r) \\
	&= B([S_{ll}]) + B([S_{lr}]) + B([S_{rl}]) + B([S_{rr}]) \\
	&+ B_{\textit{between}}(S_{ll}, S_{lr}) + B_{\textit{between}}( S_{rl}, S_{rr}) \\
	&+B_{\textit{between}}(S_l, S_r)
	=\!\sum\limits_{H\in \textit{non-leaf nodes}}\!B_{\textit{between}}(H_l, H_r)
	\end{aligned}
\end{equation}

\subsection{Metrics for interactions and the construction of a tree}

\begin{figure}[t]
	\centering
	\includegraphics[width=0.5\columnwidth]{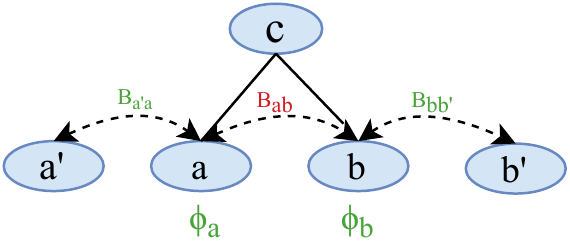}
	\caption{Interaction benefits between constituents. The interaction benefit $B_{ab}$ is more significant than $B_{a'a}$ and $B_{bb'}$,
		so the tree merges $a$ and $b$ to form a coalition $c$.}
	\label{fig:Bab}
\end{figure}

\subsubsection{Metrics for interactions.} Besides $B([S_l])$, $B([S_r])$ and $B_{\textit{between}}(S_l, S_r)$, we define three additional metrics to provide insightful analysis of interactions among words. Let us consider a sub-structure of a parent node $c$ (corresponding to the constituent $S$) and two child nodes $a$ and $b$ (corresponding to sub-constituents $S_l$ and $S_r$). As Figure~\ref{fig:Bab} shows, $a'$ is the left adjacent node of $a$, and $b'$ is the right adjacent node of $b$. We propose the metric ``\textit{density of modeled interactions}'' for a candidate coalition such as $\{a, b\}$, denoted by $r(a,b)$. This metric measures the ratio of interaction benefits between two adjacent nodes $a$ and $b$ to the total interaction benefits related to $a$ and $b$.
The density of the modeled interactions is approximated as follows.
\begin{equation}
\label{rab}
	\begin{aligned}
	r(a, b) &= \frac{\textit{interaction benefits between a and b}}{\textit{total interaction benefits related to a and b}} \\
	&\approx \frac{|B_{ab}|}{|B_{ab}| + |B_{a'a}| + |B_{bb'}| + |\phi_a| + |\phi_b|}
	\end{aligned}
\end{equation}
where { $B_{ab}=B_{\textit{between}}(S_a, S_b)$}, $\phi_a$ and $\phi_b$ can be approximated as $\phi^{(N\backslash S_a)\cup \{[S_a]\}}([S_a])$ and $\phi^{(N\backslash S_b)\cup \{[S_b]\}}([S_b])$, respectively.
To measure interaction benefits that are not represented by the tree, a metric called ``\textit{density of unmodeled interactions}" denoted by $s(a, b)$ is given.
\begin{equation}
	\begin{aligned}
	\!s(a,b) &= \frac{\textit{unmodeled interaction benefits }}{\textit{total interaction benefits related to a and b}} \\
	&\approx \frac{|B_{a'a}| + |B_{bb'}|}{|B_{ab}| + |B_{a'a}| + |B_{bb'}| + |\phi_a| + |\phi_b|}
	\end{aligned}
\end{equation}
Note that neither $r(a, b)$ nor $s(a, b)$ is an accurate estimation of the ratio of interactions. If two constituents are far away (\emph{e.g.}~not adjacent), their interaction benefits are usually small and sometimes can be neglected.
Therefore, we only consider interaction benefits between adjacent nodes (\emph{i.e.}~$B_{a'a}$, $B_{ab}$, $B_{bb'}$). We have demonstrated very little effects of such neglection in Table~\ref{ratio}. In addition,
according to Equation~\eqref{eq_intra}, we have {$B_{\textit{between}}(S_l, S_r) =  \psi^{\textit{inter}} + \psi^{\textit{intra}}_{l}+ \psi^{\textit{intra}}_{r}$}.
Therefore, we define the following metric to measure the ratio of inter-constituent interactions.
\begin{equation}
t = |\psi^{\textit{inter}}| / (|\psi^{\textit{inter}}| + |\psi^{\textit{intra}}_l + \psi^{\textit{intra}}_r|)
\end{equation}

\subsubsection{Construction of a tree.}
We use the metric $r(a,b)$ in Equation~\eqref{rab} to quantify the significance of interactions between two adjacent constituents, and to guide the construction of the tree.
We are given a trained DNN and an input sentence. The DNN can be trained for various tasks, such as sentiment classification, and the estimation of linguistic acceptability.
We construct the tree in a bottom-up manner.
Let $\Omega$ denote the set of current candidate nodes to merge.
In the beginning, each word $a_i$ of the input sentence is initialized as a leaf node, $\Omega=\{a_1, a_2,...,a_n\}$.
In each step, we compute the value of each pair of adjacent nodes $r(a_i, a_{i+1})$. Then, we select and merge two adjacent nodes with the largest value of $r(a_i, a_{i+1})$. In this way, we use a greedy strategy to build up the tree, so that salient interactions among words are represented.

\section{Experiments}

\subsubsection{Instability and accuracy of Shapley values.}
\begin{figure}[t]
	\centering
	\includegraphics[width=0.9\columnwidth]{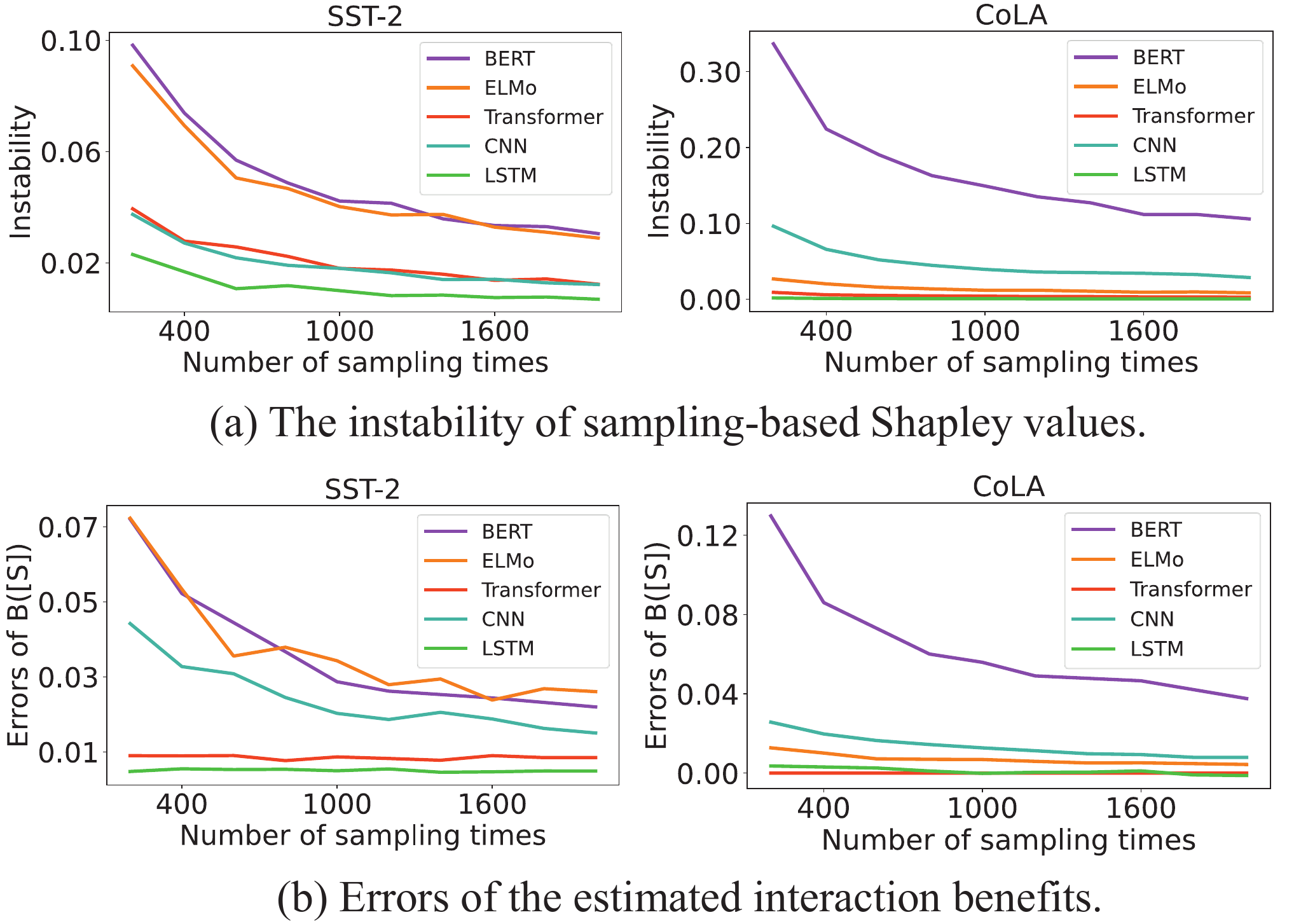}
    \caption{Evaluation of the reliability of the method.}
    \label{inst-diff}
\end{figure}

According to Equation~\eqref{shapley}, the accurate computation of Shapley values is NP-hard. \citet{castro2009polynomial} proposed a sampling-based method to approximate Shapley values with polynomial computational complexity. In order to evaluate the instability of $B([S])$, we quantified the change of the instability of Shapley values along with the increase of the number of sampling times. Let us compute the Shapley value $\phi(a)$ for each word by sampling $T$ times. We repeated such a procedure of computing Shapley values two times. Then, the instability of the computation of Shapley values was measured as $2||\boldsymbol{\phi} - \boldsymbol{\phi^{'}}|| / (||\boldsymbol{\phi}|| + ||\boldsymbol{\phi^{'}}||)$ where $\boldsymbol{\phi}$ and $\boldsymbol{\phi^{'}}$ denoted two vectors of word-wise Shapley values computed in these two times. The overall instability of Shapley values was reported as the average value of the instability of all sentences. Figure~\ref{inst-diff} (a) shows the change of the instability of Shapley values along with the number of sampling times $T$. When $T\geq 1000$, we obtained stable Shapley values.

In addition, we also evaluated the accuracy of the estimation of interaction benefits $B([S])$. The problem was that the ground truth value of $B([S])$ was computed using the NP-hard brute-force manner, according to Equation~\eqref{shapley}. Considering the NP-hard computational cost, we only conducted such evaluations on sentences with no more than 10 words. The average absolute difference (\emph{i.e.}~the error) between the estimated $B([S])$ and its ground truth value over all sentences is reported in Figure~\ref{inst-diff} (b). We found that the estimated interaction benefits were accurate enough when the number of sampling times was greater than 1000.

We found that the BERT model exhibited much higher instability and errors than other models in Figure~\ref{inst-diff}. It was because the BERT model had much stronger representation power than other models, and thus encoded more complex interactions, which was also verified in \cite{guan2019towards}. Thus, the BERT model required more sampling times.

\subsubsection{Effects of non-adjacent nodes.} To compute the density of modeled interactions $r(a, b)$, we only considered interaction benefits between two adjacent nodes, and assumed that interactions of non-adjacent nodes were much less significant than those of adjacent nodes. To verify this assumption, we defined the following metric to quantify the interaction benefit $r'(a, c)$ between two non-adjacent nodes $a$ and $c$, and evaluated whether the most salient interaction between adjacent nodes $a, b$ detected by our method was more significant than interactions between all potential non-adjacent nodes. We use $r'(a, c) = |B_{ac}| / (|B_{ac}| + |B_{a'a}| + |B_{aa''}| + |B_{c'c}| + |B_{cc''}| + |\phi_a| + |\phi_c|)$ to quantify the interaction density between non-adjacent nodes $a$ and $c$,
where $a'$ and $a''$ were the left and right adjacent nodes of $a$, $c'$ and $c''$ were the left and right adjacent nodes of $c$. If the interaction density $r(a, b)$ estimated by our method was higher than that between potential non-adjacent nodes, we considered this as a correct extraction of word interactions. Table~\ref{ratio} reports the rate of incorrect extractions of word interactions over all sentences during the construction of the tree. Based on this assumption, our method performed correctly in most cases.

\begin{table}[t]
	\centering
	\resizebox{0.75\columnwidth}{!}{
	\begin{tabular}{ccccc}
		\toprule
		\# of merges & BERT & ELMo & CNN & LSTM \\
		\midrule
		1 & 0.00 & 0.02& 0.01&  0.06\\
		\midrule
		2 & 0.00 & 0.06& 0.02&  0.13\\
		\midrule
		3 & 0.00 & 0.12& 0.02&  0.19\\
		\midrule
		4 & 0.03 & 0.15& 0.07&  0.15\\
		\midrule
		5 & 0.03 & 0.16& 0.07&  0.14\\
		\bottomrule
	\end{tabular} }
	\caption{The rate of incorrect extractions of word interactions, which verifies the assumption that effects of non-adjacent nodes can be neglected on the SST-2 dataset.}
	\label{ratio}
\end{table}

\subsubsection{Correctness of the extracted interaction.} We aimed to evaluate whether the extracted interaction objectively reflected the true interaction in the model, but the core challenge was that it was impossible to annotate ground-truth interactions between words. It was because the human's understanding of
word interactions was not necessarily equivalent to objective interactions encoded in a DNN. In this way, we conducted the following two experiments to evaluate the correctness of the extracted interactions.

\textit{Experiment 1:} In order to quantitatively evaluate the validity of the extracted interactions among words, we adopted the metric \emph{cohesion-score} proposed by \citet{chen2020generating} to justify a constituent containing significant interactions identified by our method. Given a constituent corresponding to a tree node $[p,q]=(a_p,...,a_q)$ in a sentence $x=(a_1,...,a_p,...,a_q,...,a_n)$, we picked a word in $[p,q]$ at a time, and inserted it into a random position in the sequence $(a_1,...,a_{p-1},a_{q+1},...,a_n)$. We repeated this process until there were no words left in $[p,q]$. Thus, we obtained a shuffled sentence $\tilde{x}$ from $x$. The cohesion-score measured the change of probability on the predicted class between $\tilde{x}$ and $x$ as follows:
\begin{equation}
\textit{cohesion-score}=\!\frac{1}{N}\!\sum_{i=1}^{N}\!\frac{1}{Q}\!\sum_{j=1}^{Q}(p(\hat{y_i}|{\rm x}_{i})-p(\hat{y_i}|\tilde{\rm x}_i^{(j)}))
\end{equation}
where ${\rm x}_{i}$ is the $i$-th sentence and $\hat{y_i}$ is the predicted class.
$\tilde{\rm x}_i^{(j)}$ is the $j$-th shuffled sentence from ${\rm x}_{i}$, and $Q$ was set to 100. For each sentence, we only considered the most significant constituent (\emph{i.e.}~the tree node with the maximum $\phi^{N\setminus S\cup\{[S]\}}([S])$) during the construction of the tree. We used HEDGE~\cite{chen2020generating}, which was a top-down method to recursively split a long sentence into shorter constituents, as the baseline. Besides, for a fair comparison, we reimplemented the HEDGE method to explain different NLP models trained on the SST-2 dataset. As Table~\ref{cohesion} shows, our method outperformed HEDGE, which suggested that our method extracted more significant interactions within constituents than HEDGE.

\begin{table}[t]
	\centering
	\resizebox{0.9\columnwidth}{!}{
		\begin{tabular}{lccccc}
			\toprule
			& BERT & ELMo & CNN & LSTM & Transformer \\
			\midrule
			Ours& \textbf{0.037} &\textbf{0.133} &\textbf{0.063} &\textbf{0.036} &\textbf{0.012} \\
			\midrule
			HEDGE& 0.033 &0.131&0.023 &0.004 &0.008 \\
			\bottomrule
	\end{tabular} }
	\caption{Comparisons of cohesion-scores for explanations of NLP models trained on the SST-2 dataset~\cite{socher2013recursive}.}
	\label{cohesion}
\end{table}

\begin{figure}[t]
	\centering
	\includegraphics[width=0.65\columnwidth]{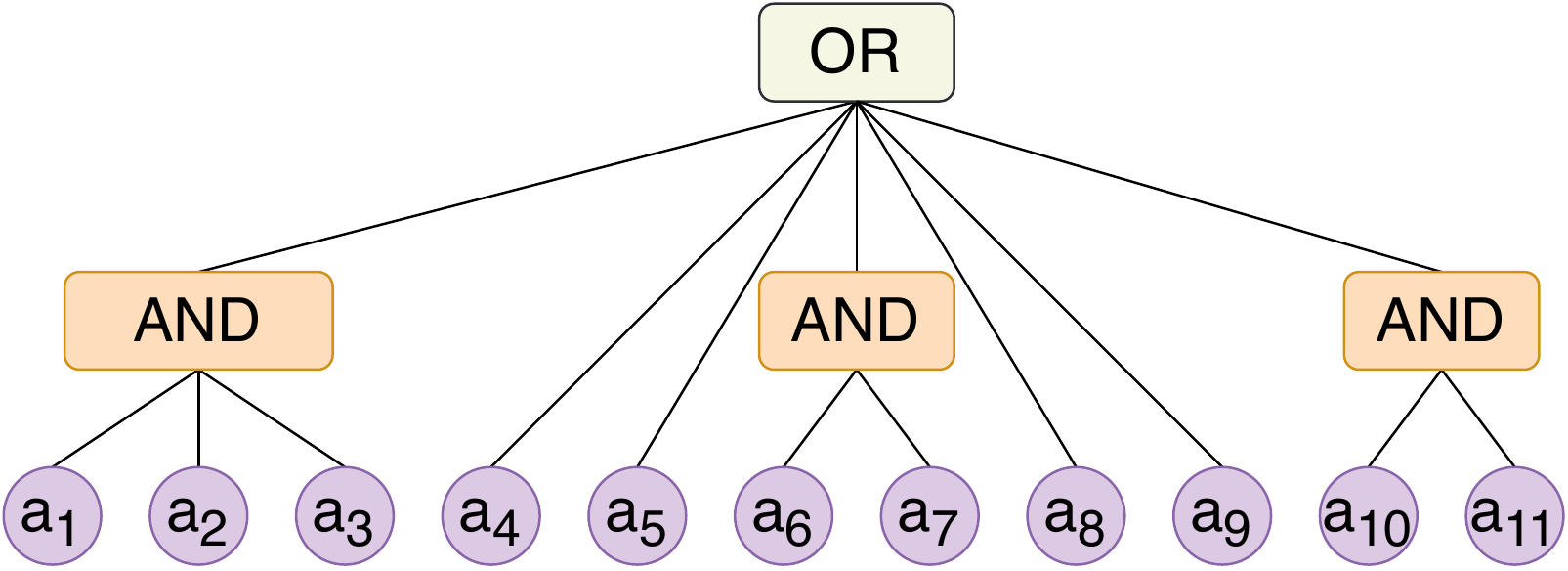}
	\caption{An example of AND-OR models. Each leaf node is a binary variable.}
	\label{fig:and-or}
\end{figure}

\renewcommand{\arraystretch}{1.2}
\begin{table}[t]
	\centering
	\resizebox{0.95\columnwidth}{!}{
		\begin{tabular}{l|cc|c|cc|c}
			\hline
			\multirow{2}{*}{} & \multicolumn{3}{c}{F1 score} & \multicolumn{3}{|c}{Recall} \\
			\cline{2-4} \cline{5-7}
			& AND-OR & OR-AND & \textbf{Avg.} & AND-OR & OR-AND & \textbf{Avg.} \\
			\hline
			Ours & 45.02 & 45.62 & \textbf{45.32} & 96.60 & 98.94 &\textbf{97.77}  \\
			\hline
			SI & 46.02 & 0.00 & 23.01  & 99.80 & 0.00 & 49.90\\
			\hline
			SI-abs & 29.77 & 29.74 & 29.76 & 61.27 & 61.22 & 61.25 \\
			\hline
			HEDGE & 46.02 & 0.00 & 23.01 & 99.80 & 0.00 & 49.90 \\
			\hline
			Random & - & - & 13.18 & - & - & 27.78 \\
			\hline
			LB & - & - & 8.35 & - & - & 18.07 \\
			\hline
			RB & - & - & 8.35 & - & - & 18.07 \\
			\hline
	\end{tabular} }
	\caption{Comparisons of the correctness of the extracted interactions on AND-OR models and OR-AND models.}
	\label{toy-example}
\end{table}

\textit{Experiment 2:} We constructed a dataset with ground-truth interactions between the inputs, as follows. The dataset was comprised of 2048 models. Each model was implemented as a boolean function, whose input was 11 binary variables $a_1, a_2,\cdots, a_{11} \in \{0, 1\}$.
The output of the model was a binary variable which consisted of AND, OR operations in a two-level tree structure (\emph{e.g.}~the tree in Figure~\ref{fig:and-or}).
More specifically, we designed 1024 models where AND operations were in the first level, and OR operations were in the second level. The other 1024 models had OR operations in the first level, and AND operations in the second level.
We evaluated whether the extracted interaction could reflect the true AND, OR constituents in the input.

The unlabeled F1 score and unlabeled recall were used to evaluate the correctness of the extracted interaction. We compared our method with six baselines.
The first baseline was \cite{lundberg2018consistent}, which defined a type of two-player interaction (\emph{i.e.}~SHAP interaction), namely \emph{SI} for short. We extended this technique to construct a tree.
\emph{I.e.}~we recursively merged the two adjacent nodes with the largest SHAP interaction value.
The second baseline was similar to the first one. This baseline used the largest absolute SHAP interaction value (\emph{i.e.}~the significance) to construct the tree, namely \emph{SI-abs}.
The third baseline was HEDGE~\cite{chen2020generating}, as mentioned above.
Since there was no other method to construct a tree for interactons to the best of our knowledge, the other three baselines Random, left-branching (LB) and right-branching (RB) trees (used by \citet{shen2018neural} as baselines) were selected as trivial solutions.
As Table~\ref{toy-example} shows, our method outperformed all baselines.
Note that theoretically, there did not exist a 100\% F1 score, because the extracted binary tree was naturally different from the ground-truth n-ary tree.

\begin{figure}[t]
	\centering
	\includegraphics[width=0.75\columnwidth]{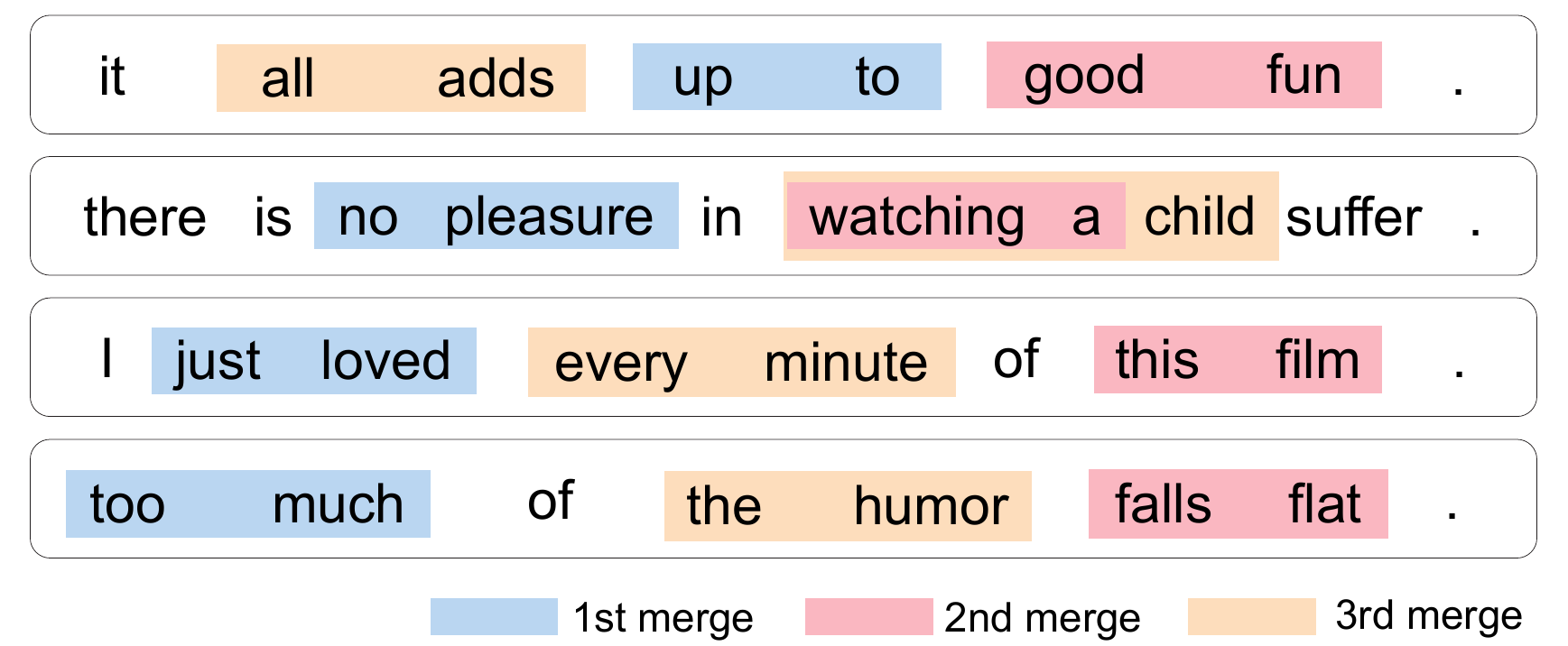}
	\captionof{figure}{Examples of the phenomenon that constituents with distinct emotional attitudes have strong interactions and are extracted in the first three steps for BERT learned on the SST-2 dataset.
	}
	\label{sst-example}
\end{figure}

\begin{figure}[t]
	\centering
	\includegraphics[width=0.9\columnwidth]{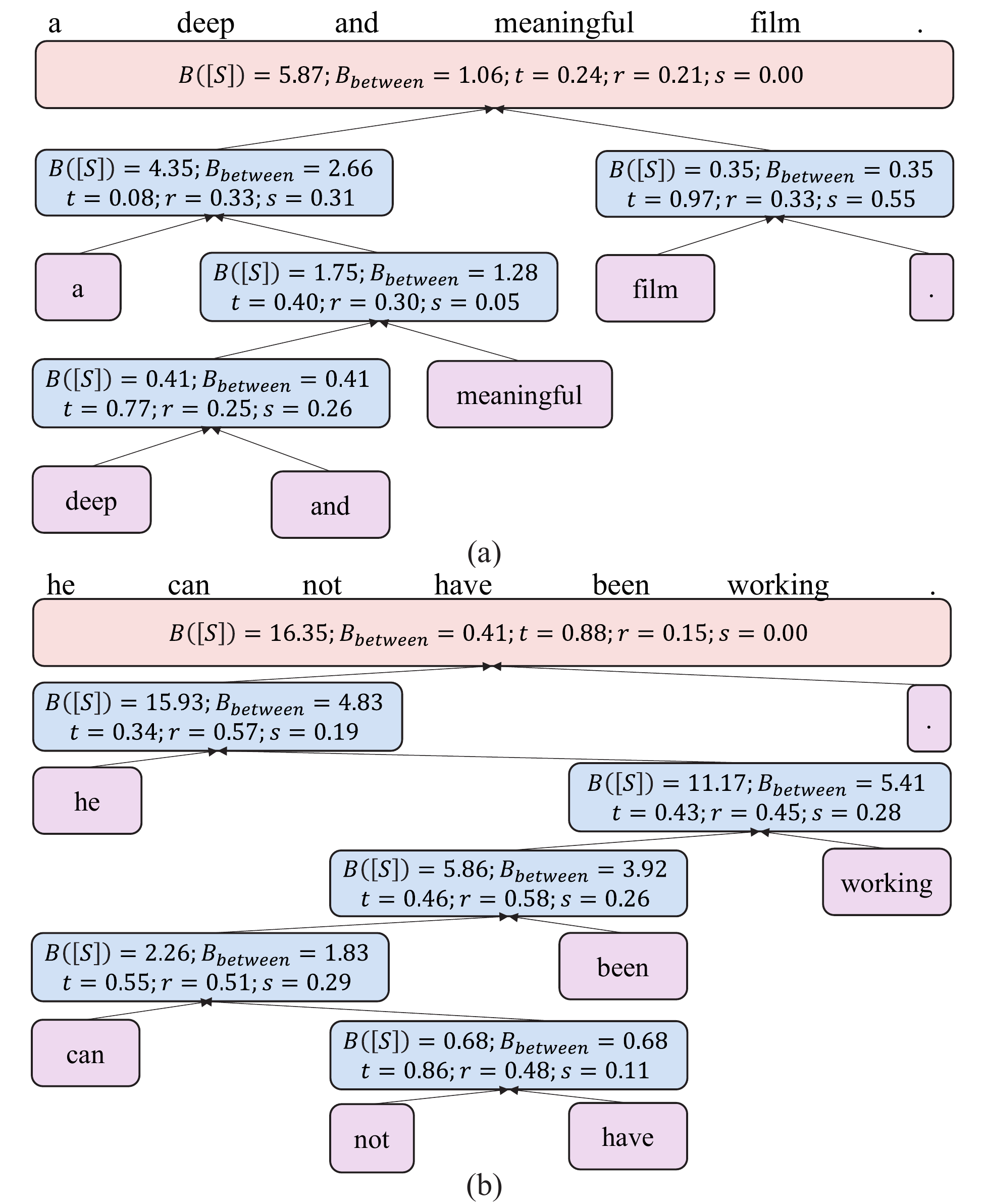}
	\caption{Examples of trees extracted from BERT trained on the SST-2 dataset (a) and the CoLA dataset (b), respectively. Metrics are shown in each non-leaf node.
	}
	\label{fig:bert_trees}
\end{figure}

\subsubsection{Comparisons of trees generated by different DNNs.} We learned DNNs for binary sentiment classification based on the SST-2 dataset~\cite{socher2013recursive}, and learned DNNs to predict whether a sentence was linguistically acceptable based on the CoLA dataset~\cite{warstadt2018neural}. For each task, we learned five DNNs, including the BERT~\cite{devlin2018bert}, the ELMo~\cite{peters2018deep}, the CNN proposed in \cite{kim2014convolutional}, the two-layer unidirectional LSTM~\cite{hochreiter1997long}, and the Transformer~\cite{vaswani2017attention}.

We used our method to extract tree structures that encoded interactions among words inside various trained DNNs. Figure~\ref{fig:bert_trees} illustrates trees extracted from BERT on different tasks. \textbf{(1)}
For the sentiment analysis task, as Figure~\ref{sst-example} shows, most trees of these DNNs usually extracted constituents with distinct positive/negative emotional attitudes in early stages. \textbf{(2)} For the linguistic acceptability task, BERT usually combined noun phrases firstly, while the subject was combined almost at last. CNN was prone to construct a tree with a ``subject+verb-phrase+noun/adjective-phrase'' structure. ELMo and LSTM usually extracted small constituents including a preposition or an article, \emph{e.g.}~``vacation in,'' ``the earth.'' Transformer tended to encode interactions among adjacent constituents sequentially.

\begin{figure}[t]
	\centering
	\includegraphics[width=0.9\columnwidth]{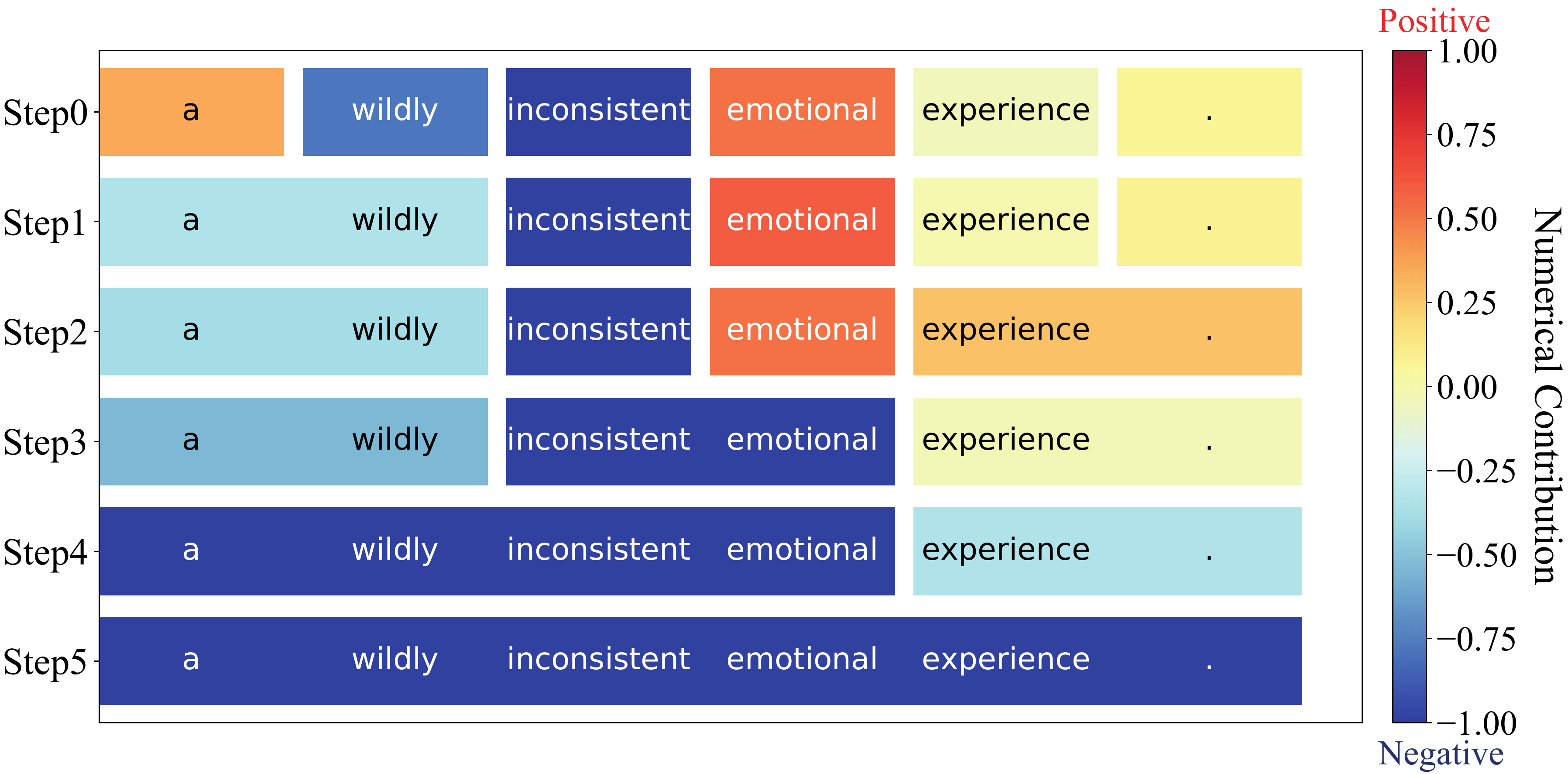}
	\caption{Effects of the extracted interactions. The extracted interactions significantly affected the contributions of constituents. For example, significant interactions between ``inconsistent" and ``emotional" made the positive word ``emotional" negative, which eventually guided the DNN to make the correct prediction.}
	\label{decision-making}
\end{figure}

\emph{Analysis of significant interactions reflected by the tree:} To understand how interactions among words affected the DNN to make a decision, we quantified the contribution of each word/constituent $\phi^{N\setminus S\cup\{[S]\}}([S])$ (\emph{i.e.}~$\phi_a$ in Equation~\eqref{rab}) to the model prediction with sampling times $T=2000$ during the construction of the tree.
As Figure~\ref{decision-making} shows, the DNN encoded significant interactions
(\emph{inconsistent}, \emph{emotional}), (\emph{a wildly}, \emph{inconsistent emotional}), etc. to correctly predict the whole sentence as negative. During the construction of the tree, we discovered how words/constituents interacted to affect the model prediction. This provided a better understanding of the logic encoded in the DNN.

\begin{table}[t]
	\centering
	\resizebox{0.8\columnwidth}{!}{
		\begin{tabular}{lcccccccc}
			\toprule
			Dataset & BERT & ELMo & CNN & LSTM \\
			\midrule
			CoLA  &36.06 & 15.38 & 15.19 & 12.65\\
			\midrule
			SST-2 &17.67 & 16.72 & 11.69 & 29.06 \\
			\midrule
			& Transformer & Random & LB & RB \\
			\midrule
			CoLA & 3.45 &15.12 &2.68 & 60.46 \\
			\midrule
			SST-2 & 23.49 &16.32& 12.27& 47.35 \\
			\bottomrule
	\end{tabular} }
	\caption{Fitness (the unlabeled F1 score) between the extracted trees from NLP models and syntactic trees, which demonstrates that interactions encoded in a DNN are not quite related to the syntactic structure.}
	\label{distance_res}
\end{table}

\emph{Comparisons of the fitness between the extracted trees and syntactic trees: }
Furthermore, we compared the fitness between the automatically extracted tree and the syntactic tree of the sentence.
To this end, given an
input sentence, we used the Berkeley Neural Parser~\cite{Kitaev-2018-SelfAttentive}
to generate the syntactic tree as the ground-truth.\footnote{The parser's performance was good enough to take its parsing results as ground-truth.} We used the unlabeled F1 score to evaluate the fitness. Experimental results are reported in Table~\ref{distance_res}, which demonstrates the logic of interactions modeled by the DNN was significantly different from human knowledge.

\textbf{In addition, our method can also be applied to build a tree for interactions \emph{w.r.t.} the computation of features in an intermediate layer.}

\section{Conclusion}

In this paper, we have defined and extracted interaction benefits among words encoded in a DNN, and have used a tree structure to organize word interactions hierarchically. Besides, six metrics are defined to disentangle and quantify interactions among words. Our method can be regarded as a generic tool to objectively diagnose various DNNs for NLP tasks, which provides new insights of these DNNs.

%



\section{Acknowledgments}
This work is partially supported by National Natural Science Foundation of China (61906120 and U19B2043).

\section*{Ethics Statement}
This study has broad impacts on the understanding of signal processing in DNNs for NLP tasks. Our work provides researchers in the field of explainable AI with a generic tool to quantify the inter-word interactions encoded by a trained DNN. Currently, existing methods mainly focus on interactions between two variables or two words. Our research proposes new metrics to quantify interactions among multiple variables and develops a method to build up a tree representing the hierarchical structures of interactions. As a generic tool to analyze DNNs, we have applied our method to classic DNNs and have obtained several new insights on signal processing encoded in DNNs for NLP tasks.

\bibliography{refpaper}

\end{document}